\documentclass[10pt]{wlscirep}
\usepackage[utf8]{inputenc}
\usepackage[T1]{fontenc}
\title{P-ODN: Prototype based Open Deep Network for Open Set Recognition}

\author[1,3]{Yu Shu}
\author[2,3]{Yemin Shi}
\author[4]{Yaowei Wang}
\author[2,3]{Tiejun Huang}
\author[2,3,4]{Yonghong Tian\thanks{Correspondence should be addressed to Yonghong Tian (email: yhtian@pku.edu.cn).}}
\affil[1]{School of AAIS, Peking University, Beijing, China}
\affil[2]{School of EE\&CS, Peking University, Beijing, China}
\affil[3]{National Engineering Laboratory for Video Technology, China}
\affil[4]{Peng Cheng Laboratory, China}

\begin{abstract}
Most of the existing recognition algorithms are proposed for closed set scenarios, where all categories are known beforehand. However, in practice, recognition is essentially an \textit{open set} problem. There are categories we know called ``knowns'', and there are more we do not know called ``unknowns". Enumerating all categories beforehand is never possible, consequently it is infeasible to prepare sufficient training samples for those unknowns. Applying closed set recognition methods will naturally lead to unseen-category errors. To address this problem, we propose the prototype based Open Deep Network (P-ODN) for open set recognition tasks. Specifically, we introduce prototype learning into open set recognition. Prototypes and prototype radiuses are trained jointly to guide a CNN network to derive more discriminative features. Then P-ODN detects the unknowns by applying a multi-class triplet thresholding method based on the distance metric between features and prototypes. Manual labeling the unknowns which are detected in the previous process as new categories. Predictors for new categories are added to the classification layer to ``open" the deep neural networks to incorporate new categories dynamically. The weights of new predictors are initialized exquisitely by applying a distances based algorithm to transfer the learned knowledge. Consequently, this initialization method speed up the fine-tuning process and reduce the samples needed to train new predictors. Extensive experiments show that P-ODN can effectively detect unknowns and needs only few samples with human intervention to recognize a new category. In the real world scenarios, our method achieves state-of-the-art performance on the UCF11, UCF50, UCF101 and HMDB51 datasets.
\end{abstract}
\begin{document}

\flushbottom
\maketitle
%
%
\thispagestyle{empty}

\section*{Introduction}

Deep neural networks have demonstrated significant performance on many visual recognition tasks \cite{krizhevsky2012imagenet,wang2016temporal,shi2017learning}. Almost all of them are proposed for closed set scenarios, where all categories are known beforehand. However, in practice, some categories can be known beforehand, but more categories can not be known until we have seen them. We call the categories we know as priori the ``knowns'' and those we do not know beforehand the ``unknowns''. Enumerating all categories is never possible for the incomplete knowledge of categories. And preparing sufficient training samples for all categories beforehand is time and resource consuming, which is also infeasible for unknowns. Consequently, applying closed set recognition methods in real scenarios naturally leads to unseen-category errors. Therefore, recognition in the real world is essentially an open set problem, and an open set method is more desirable for recognition tasks.

\begin{figure*}
	\centering{\includegraphics[width=5in]{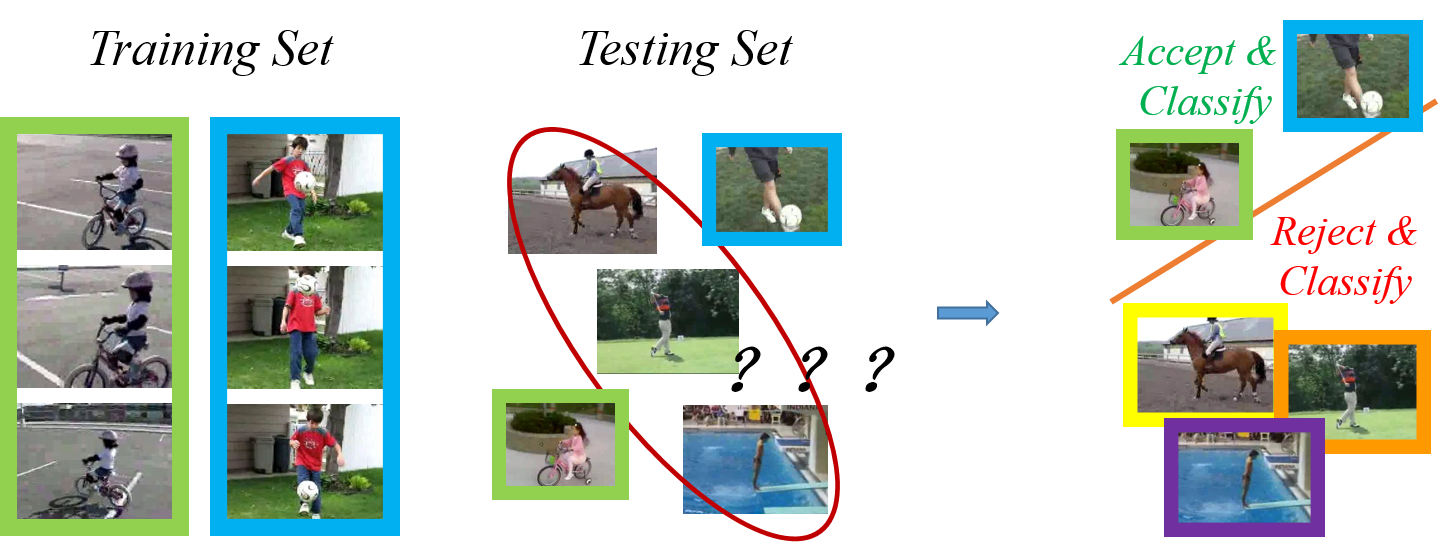}}
	\caption{Open Set Recognition. The training set contains sufficient labeled samples of different categories (colored in green and blue), while the testing set contains knowns but also unknowns which haven't been seen at all. The solution should be able to accept the knowns and reject the unknowns. Simultaneously, the solution should classify knowns into correct known categories and further be able to classify unknowns as well.}
	\label{motivation}
\end{figure*}

As shown in Fig. \ref{motivation}, in open set recognition problem, categories of the training set have sufficient labeled samples, which are knowns, while the testing set contains both knowns and unknowns. The solution to the problem should be able to accept and classify knowns into correct known categories and also reject unknowns. Simultaneously, it is natural to further extend to perform recognition of unknowns as shown in the right part of Fig. \ref{motivation} with different color boxes classifying the unknowns.

Technically speaking, many methods on incremental learning can be used to handle new instances of known categories \cite{tveit2003multicategory, cauwenberghs2001incremental, crammer2006online, yeh2008dynamic, herbster2001learning}. However, most of these approaches do not consider about unknowns or dynamically adding new categories to the system. In \cite{mensink2012metric}, a discriminative metric is learned for Nearest Class Mean (NCM) classification on the knowns, and new categories are added according to the mean features. This approach, however, assumes that the number of known categories is relatively large. An alternative multi-class incremental approach based on least-squares SVM has been proposed by Kuzborskij \textit{et al.} \cite{kuzborskij2013n} where for each category a decision hyperplane is learned. However, in this way, every time a category is added, the whole set of the hyperplanes will be updated again, which is too expensive as the number of categories growing. 

In particular, most of the researches on open set recognition focus on detecting unknowns only, recent works \cite{Scheirer2013Transactions, Scheirer2014Probability, bendale2016towards} have established formulations of classifying knowns and rejecting unknowns. While it is natural to extend to classify unknown samples after detecting unknowns. And in the real world, solutions which further classify the unknowns are more challenging and have a wider range of applications. Abhijit \textit{et al.} \cite{bendale2015towards} proposed a SVM-based recognition system that could continuously recognize new categories in an open world model by extending the NCM-like algorithms \cite{ristin2014incremental} to a Nearest Non-Outlier (NNO) algorithm. But it is not applicable in deep neural networks, and the performance is much worse than deep neural network based algorithms. Recently, in the work \cite{yang2018robust}, Yang \textit{et al.} have tried to handle the open set recognition problem by training prototypes to represent the unknowns. But the solution works on the assumption that all unknowns have sufficient labeled samples to train discriminative prototypes, which is not realistic. And the system needs to be retrained when new categories comes, which is time and computational resource consuming.

In our previous work \cite{shuyu2018odn}, we proposed an Open Deep Network (ODN) algorithm for open set recognition. First, we train a CNN network to classify the knowns which have sufficient samples. Then the triplet threshold of each category is calculated based on the correctly classified features of training set. Unknowns can be detected by applying the triplet thresholds on the features derived by the CNN. Manual labeling the unknowns which are detected in the previous process, and predictors of the classification layer are added dynamically to incorporate new categories. Weights of the new predictors are initialized by applying the emphasis initialization method which transfers the learned knowledge of the CNN to speed up the fine-tuning. However, the triplet thresholds are calculated on the sampled features of training set, consequently, unknowns detection process might be affected by the outliers of training set. Besides, relations of categories are defined on the feature scores in emphasis initialization method, which is a simple way to estimate the similarity of categories.


Most recently, the prototype learning was introduced to improve the robustness of CNNs. Yang \textit{et al.} proposed the CPL to improve the robustness by using prototypes and proposed the PL (prototype loss) to improve the intra-class compactness and inter-class distance of the feature representation in the work \cite{yang2018robust}. Yang \textit{et al.} also introduced a method to handle open set recognition problem by using prototypes in their paper. However,  as mentioned before, this method assumes that samples of unknowns are sufficient to train the prototypes. And when new unknowns come, the system needs to be retrained again. Inspired by the prototype learning concept, we propose the prototype based Open Deep Network (P-ODN) to handle the open set recognition problem.

In this paper, we propose P-ODN to improve the robustness in detecting unknowns and updating deep neural networks, consequently facilitating open set recognition. Basically, prototypes and prototype radiuses are trained jointly to derive more precise features to better represent categories. In \textit{prototype module}, prototypes are taught to learn the centers of knowns. And in \textit{prototype radius module}, values of prototypes are further restricted to a curtain range by learning a radius for each category as a regularization item of prototypes. Both of the modules help to improve the intra-class compactness and inter-class distance of the feature representation. Then the correctly classified features of training set are projected into a different feature space by calculating the distance distribution between the features and prototypes. The triplet thresholds are learned based on the correctly classified distance distribution. Instead of detecting unknowns directly on the features, based on the statistic information of training samples, detecting unknowns based on the distance of prototypes keeps knowledge of the model, which has less potential to be effected by the outliers of training set. After manual labeling the unknowns which are detected in the previous process, new predictors are initialized based on the distance distribution of new samples and prototypes. Each weight column of the knowns is integrated to initialize the new weight according to the distance distribution. And the distance distribution contains more robust relation knowledge of the new category and knowns. Finally, fine-tuning the model with the manual labeled samples to incorporate new categories. 

In order to give a convincing results of our P-ODN, in this paper, we choose to focus on the action recognition problem which is a more challenging recognition task. The effectiveness of the proposed framework is evaluated on four public datasets: UCF11, UCF50, UCF101 and HMDB51. The experimental results show that our method can effectively detect unknowns and needs only few samples with human intervention to recognize a new category. And our method achieves the state-of-art performance on all the four datasets in real world scenarios.

\begin{figure*}
	\centering{\includegraphics[width=6in]{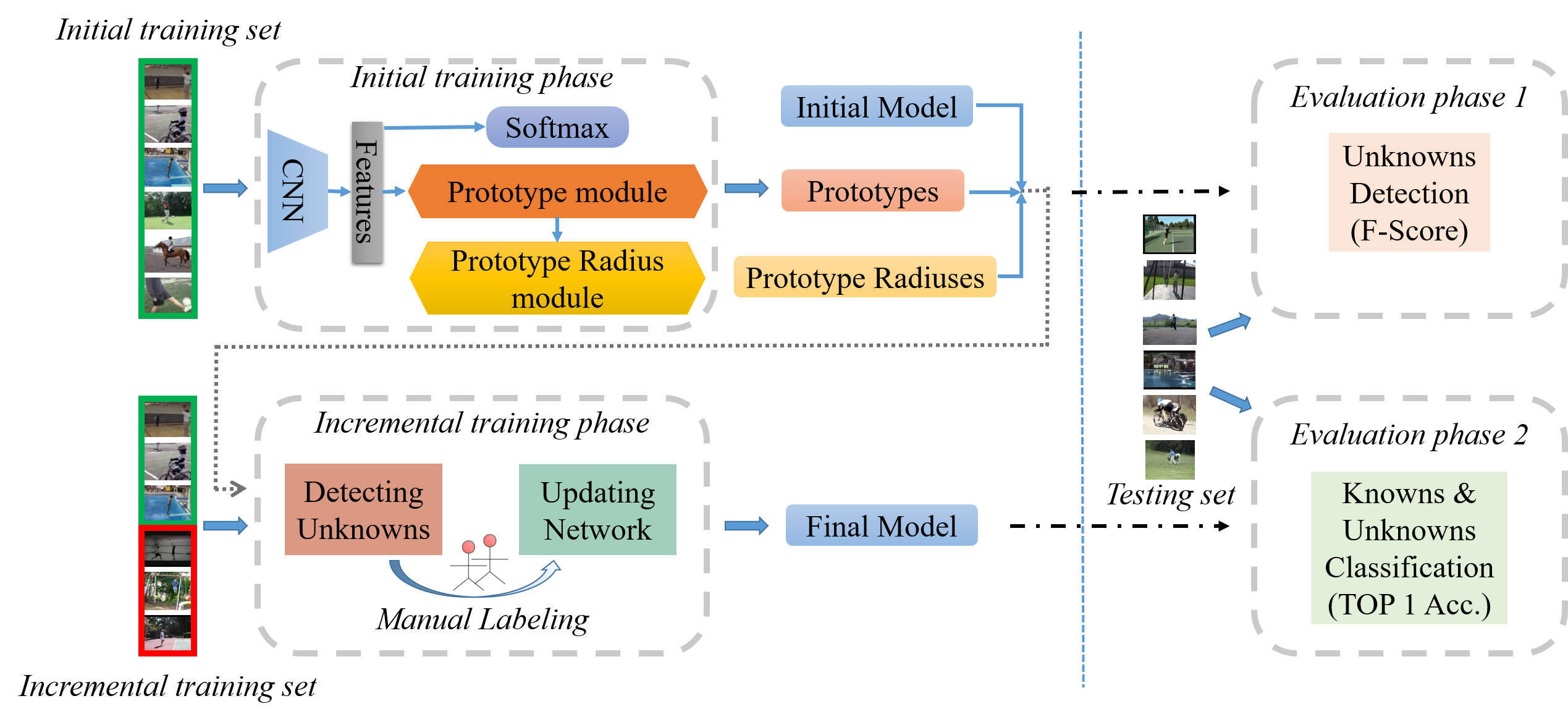}}
	\caption{Framework of open set recognition. The left part of the blue dotted line illustrates the two training phases: a) Initial training phase takes the initial training set (contains the knowns only) as input, then learns and outputs an initial model, prototypes and prototype radiuses for each category. b) Incremental training phase takes the incremental training set (contains both knowns and unknowns) and the outputs of Initial training phase as inputs, then detects the unknowns. Manual labeling the unknowns which are detected in the previous process. Next, the new category is dynamically incorporated in the model. Finally, fine-tuning the model with only few samples to make the unknowns ”known”. The right part of the dotted line illustrates two evaluation phases responding to the two training phases: a) Evaluation phase 1, the detection f-score of unknowns is measured here on the initial model trained in Initial training phase. b) Evaluation phase 2, the classification accuracy of both knowns and unknowns is measured here on the final model trained in Incremental training phase.}
	\label{framework}
\end{figure*}

\section*{Overview} \label{overview}
The framework of our open set recognition approach is shown in Fig. \ref{framework}. Two training phases and two evaluation phases constitute the whole framework. The initial training set which contains only knowns is provided to the Initial training phase as input. Then an initial model is trained, as well prototypes and prototype radiuses of categories. The Incremental training phase takes the incremental training set (contains both knowns and unknowns) as input and extracts the features by using the initial model. Then distances of the features and the prototypes are measured under the constraint of the prototype radiuses. Next, a triplet thresholding method proposed in our previous work \cite{shuyu2018odn} is modified to apply in our framework to detect the unknowns. Manual labeling the unknowns as new categories which are dynamically incorporated in the model. Then fine-tuning the model with only few samples to make the unknowns ”known”. Finally, the final output model can classify both knowns and unknowns. 

Two evaluation phases are set to evaluate the model performance. Evaluation phase 1 is carried out after the Initial training phase. Testing set which contains both knowns and unknowns is provided to the initial model, and we measure the detection f-score of unknowns at this phase. Evaluation phase 2 is carried out after the Incremental training phase. Classification TOP 1 accuracy of both knowns and unknowns is measured here as the most important performance indicator of open set recognition tasks.

\section*{Prototype based Open Deep Network}\label{sec:podn}

\begin{figure*}
	\centering{\includegraphics[width=6in]{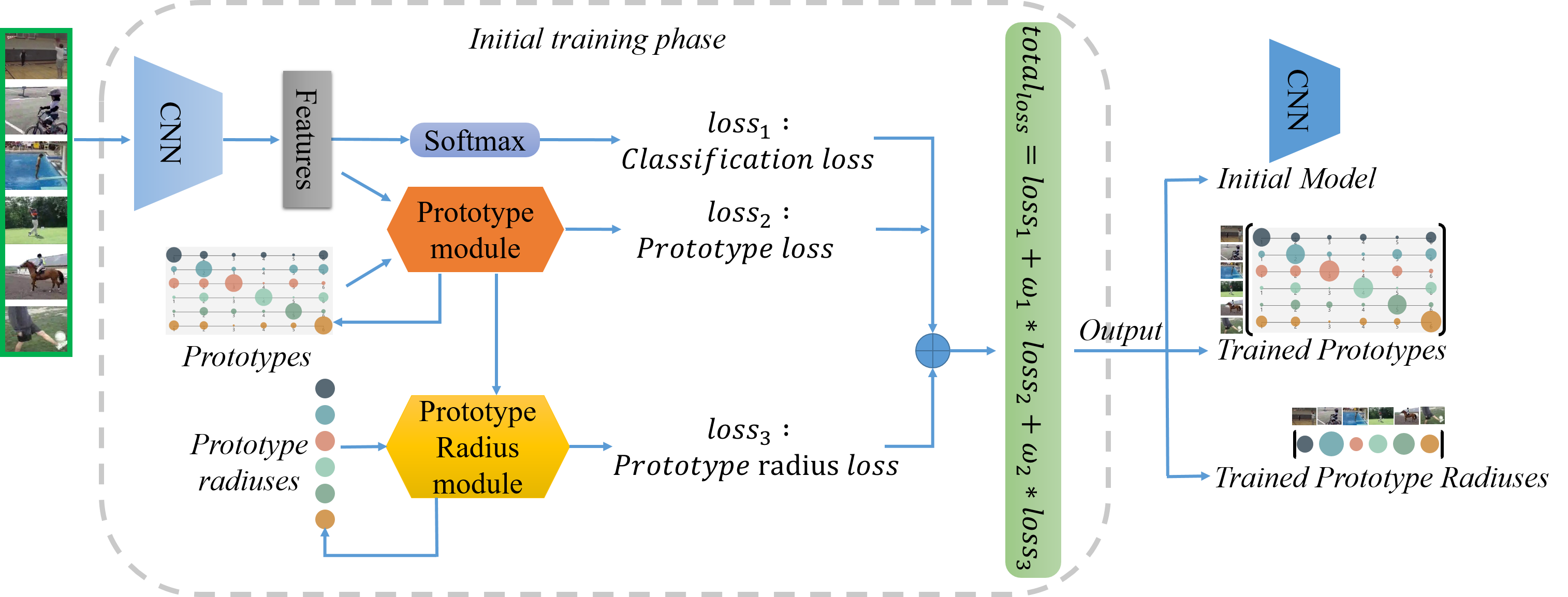}}
	\caption{Structure of prototype based open deep network (P-ODN) in the Initial training phase. The CNN takes knowns as input, and a classification loss is applied to train the initial classification model. Two major modules: a) Prototype module takes the features extracted by CNN as input and learns prototypes for categories. b) Prototype Radius module takes the prototype based distances as input, where the distances are calculated in the Prototype module, and learns the scope of each category prototype. Finally, the Initial training phase outputs an initial model, trained prototypes and prototype radiuses for the Incremental training phase later.}
	\label{main_method}
\end{figure*}

The structure of prototype based open deep network (P-ODN) in the Initial training phase is shown in Fig. \ref{main_method}. Basically, this phase includes two major modules: first, a \textit{Prototype module} is applied to learn prototypes of categories based on the prototype learning. Second, in order to guarantee each prototype of the category in a certain range, a \textit{Prototype Radius module} is proposed. Each category will learn a prototype radius to further restrict the scope of features derived by the model. Three kinds of losses are applied to train prototypes and prototype radiuses. First we apply the cross entropy loss to train the classification capacity of the neural networks, which we denotes as $loss_{1}$:

\begin{equation}\label{loss1}
loss_{1} = -\frac{1}{S}\sum_{i=1}^{S}[label_{i}*log(softmax(f_{i}))]
\end{equation}
where $S$ is the batch size, $f_{i}$ is the feature of the $i$th sample in the batch, and $label_{i}$ is the ground truth. Second, the prototype loss, which is firstly proposed by \cite{yang2018robust}, is modified to apply in our framework to train the prototypes of knowns. Third, we propose the prototype radius loss, which guides the model to learn the radius scope of each known category. 


The Initial training phase outputs the initial model, trained prototypes and prototype radiuses, and they will be used later in the Incremental training phase. 

Major modules (\textit{Detecting Unknowns} and \textit{Updating Network}) of P-ODN in the Incremental training phase are shown in Fig. \ref{framework}. The initial model trained in the Initial training phase extracts the features of the incremental training set here. We will simply review a triplet thresholding method of unknowns detection. And the method is modified to be applicable in the P-ODN to detect the unknowns based on the distance metric of the features and the trained prototypes. Then, a new distances based weights initialization method is introduced to initialize the weights of new category predictors in the \textit{Updating Network module}. After the initialization of new weights, few manual labeled samples are used to fine-tune the model. New categories are incorporated in the current model continuously. At the end of this phase, a final model which can handle both knowns and unknowns is trained. 

\subsection*{Prototype module}\label{subsec:pm}
\begin{figure*}
	\centering{\includegraphics[width=6in]{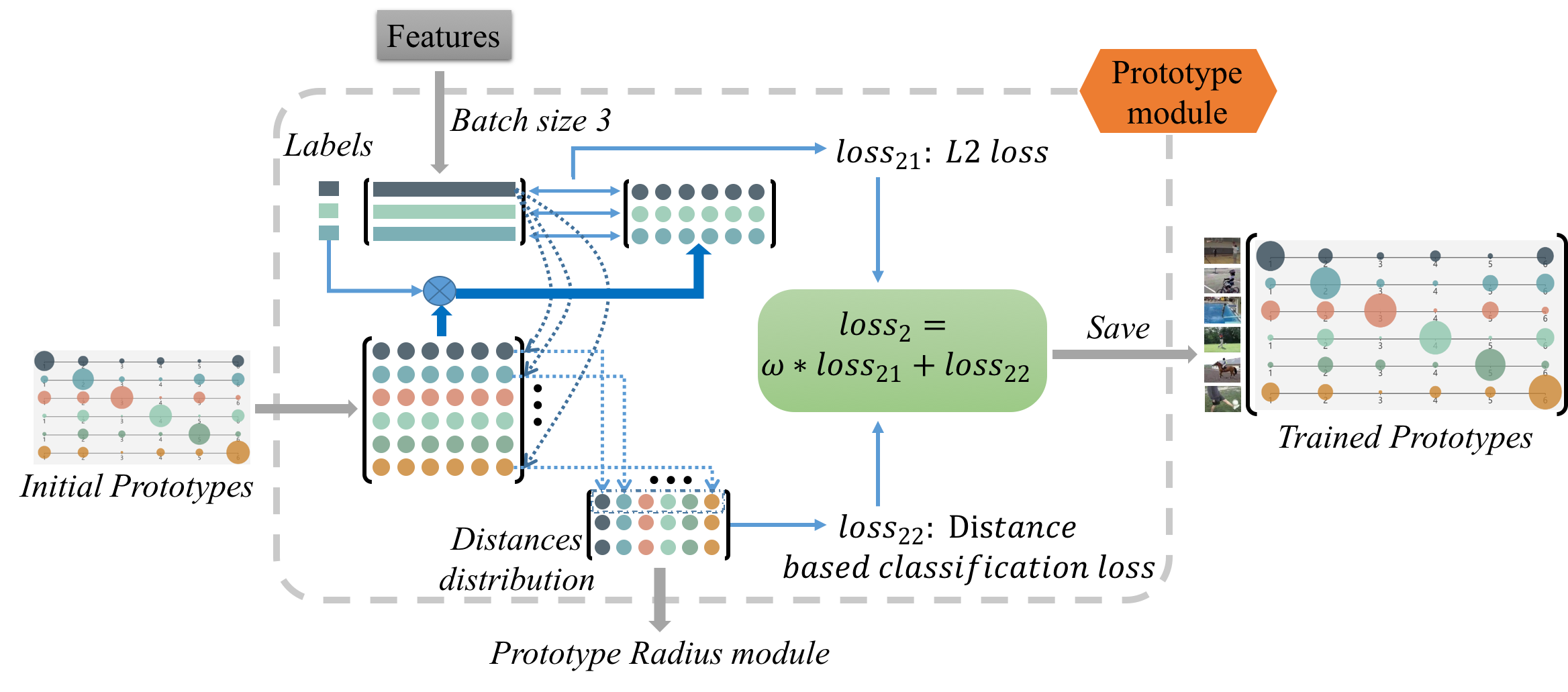}}
	\caption{The illustration of Prototype Module. Different colors represent different categories in the figure. To give an explicit explanation of the process, we assume the batch size is 3 as shown in the figure. The corresponding prototypes of categories are chosen according to the labels of features. Then a L2 loss is applied to indicate the prototypes to learn. Simultaneously, a distance distribution matrix is calculated to train the capacity of the prototypes with the distance based classification loss.}
	\label{prototype_module}
\end{figure*}

Fig. \ref{prototype_module} illustrates the algorithm of training the prototypes to represent the centers of knowns. Since prototype learning has shown its effectiveness in increasing the inter-class variation \cite{yang2018robust}, we introduce the prototype learning into open set recognition tasks and further use prototypes to detect unknowns.

Specifically, a $N \times N$ prototype matrix is initialized with zeros, where N is the category number of knowns. Each row of the prototype matrix, shown in different color in Fig. \ref{prototype_module}, represents the prototype (or center) of each known category. The prototype loss ($loss_{2}$) is applied to acquire the trained prototypes which vary greatly in different categories. The prototype loss consist of a L2 loss ($loss_{21}$) and a distance based classification loss ($loss_{22}$). Then the two losses are combined with a weight argument $\omega$:

\begin{equation}\label{loss2}
loss_{2} = \omega * loss_{21} + loss_{22}
\end{equation}

\textbf{The L2 loss}: $S$ prototypes are chosen according to the label of features, where $S$ is the batch size of the CNN networks. To give an explicit explanation of the process, we assume the batch size is 3 as shown in Fig. \ref{prototype_module}. The L2 loss is applied on the chosen prototypes and the features to guide prototypes to learn the characters of the features:

\begin{equation}\label{L2_loss}
loss_{21} = -\frac{1}{2S}\sum_{i = 1}^{S}(f_{i} - p_{i})^{2}
\end{equation}
where $f_{i}$ is the feature of the $i$th data sample in the batch and $p_{i}$ is the corresponding prototype.

\textbf{The distance based classification loss}: As the prototypes and features are trained jointly, simply applying the L2 loss to make prototypes be similar to the features would be instable. The prototypes would be easily misled by some outliers of the training data samples. We add the distance based classification loss to improve the classification capacity of the prototypes and increasing the penalty of misclassification samples, which helps to learn more stable and characteristic prototypes of categories.

As shown in Fig. \ref{prototype_module}, the Euclidean distance of each feature and each category prototype is calculated to get a distance distribution matrix $D$:

\begin{equation}\label{dc_loss}
D_{ij} = \frac{1}{||f_{i} - p_{j}||_{2}^{2} + \varepsilon}
\end{equation}
where $i = \{1, \cdots, S\}$ and $j = \{1, \cdots, N\}$. We take the reciprocal of distances between features and prototypes here to make features near to the prototypes get larger probability value. And $\varepsilon = 0.001$ is applied to avoid dividing by zero. So classification can be implemented by assigning label according to the largest value in each row of $D$. Then the cross entropy loss is applied on $D$, the $loss_{22}$:

\begin{equation}\label{loss_{22}}
loss_{22} = -\frac{1}{S}\sum_{i=1}^{S}[label_{i}*log(softmax(D[i,:]))]
\end{equation}
where the $S$ is the batch size (the same as the row number of $D$), $label_{i}$ is the ground truth, and $D[i,:]$ denotes the $i$th row of $D$.

In this module, P-ODN learns the category prototypes by applying the $loss_2$, then the trained prototypes are saved which will be used in the Incremental training phase.

\subsection*{Prototype Radius module}\label{subsec:prm}
The \textit{prototype radius module} is a key part of the initial training phase which aims to restrict the values of prototypes to a certain range and learn the prototype radius of categories. The prototypes and the prototype radiuses are trained jointly by adding a L2 loss ($loss_{3}$) which can be regard as a regularization item of the prototype learning.

\begin{figure*}
	\centering{\includegraphics[width=6in]{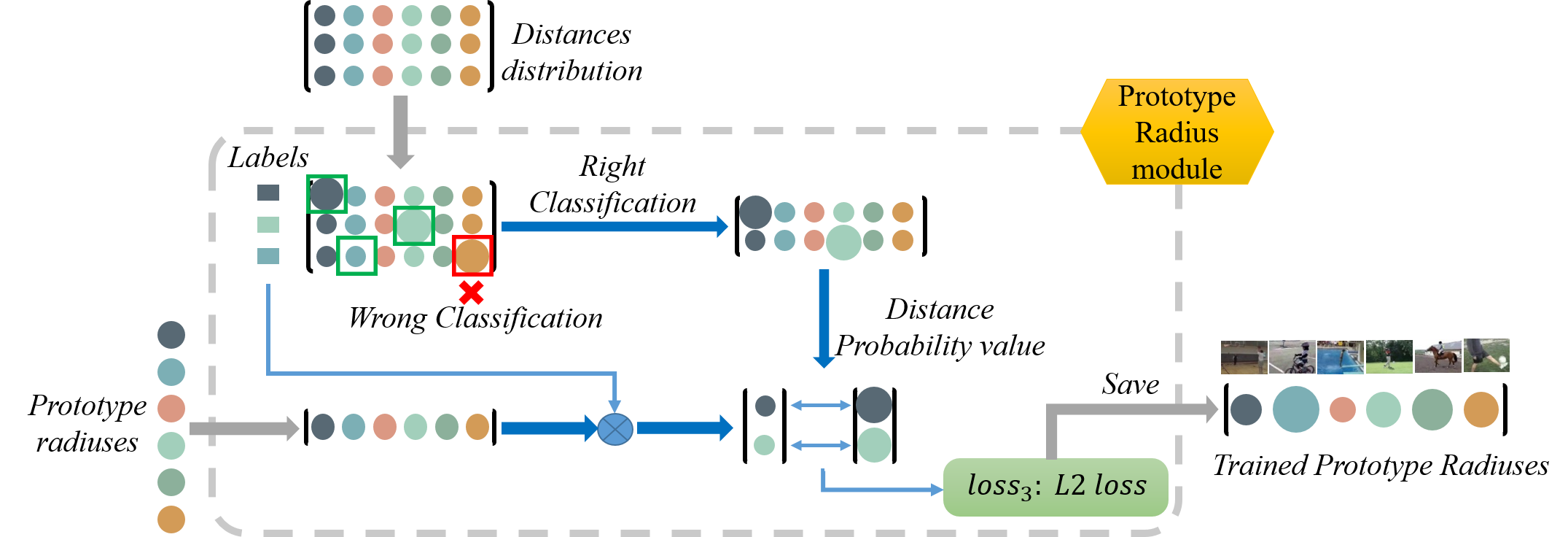}}
	\caption{The illustration of Prototype Radius Module. To restrict the features to a curtain range, the prototype radius module is applied. First, the correctly classified rows of the distance distribution matrix are chosen. Then the distance values of the corresponding categories are used to train the prototype radiuses with a L2 loss. This module can be regarded a regularization item of the prototypes.}
	\label{prototype_radius_module}
\end{figure*}

As shown in Fig. \ref{prototype_radius_module}, a vector of each category prototype radius is initialized with zeros. The distance distribution matrix ($D$) calculated in the \textit{prototype module} is inputed to the \textit{prototype radius module}.  Then the correctly classified probability scores of $D$ are chosen to guide the prototype radiuses to learn, the $loss_{3}$:

\begin{equation}
loss_{3} = -\frac{1}{2T}\sum_{t=1}^{T}(r_{t} - d_{t})^{2}
\end{equation}
where $T$ is the number of correctly classified distance probability values, $d_{t}$ is the $t$th correctly classified distance probability value (the largest score of distance distribution row), and $r_{t}$ is the category prototype radius which is chosen according to the labels of samples.

The prototype radiuses act like a memory unit of the distance probability values of the correctly classified samples. It is updated according to the correctly classified samples continuously. Simultaneously it stores the distance probability value information of the previous correctly classified samples. So the \textit{prototype radius module} integrates the long-term information to restrict the distance distribution value to a certain range and indirectly restricts the scope of the features extracted by the model. The use of the long-term information improves a lot especially in the temporal stream of action recognition.

In this module, P-ODN learns the category prototype radiuses by applying the $loss_3$ and the trained prototype radiuses are also saved. Note that the total loss of the initial training phase goes as:
\begin{equation}
total\_loss = loss_{1} + w_{1}*loss_{2} + w_{2}*loss_{3}
\end{equation}
where $w_{1}$ and $w_{2}$ are weight arguments, we set as $0.1$ and $0.01$ in our experiments.

Obviously, the prototype radiuses are trained jointly with the prototypes, which play an important role as a regularization item to train the prototypes.


\subsection*{Detecting Unknowns}\label{subsec:ud}

In our previous work \cite{shuyu2018odn}, we proposed a multi-class triplet thresholding method to detect the unknowns. Basically, a triplet threshold ($[\eta, \mu, \delta ]$) per category is calculated, i.e. accept threshold $\eta$, reject threshold $\mu$ and distance-reject threshold $\delta$.The triplet threshold $[ \eta_{i}, \mu_{i}, \delta_{i} ]$  of category $i$ is calculated as 
\begin{align}
\eta_{i} =& \frac{1}{X_{i}}\sum_{j=1}^{X_{i}}F_{i,j}\\
\mu_{i} =& \varepsilon * \eta_{i}\\
\delta_{i} =& \rho * \frac{1}{X_{i}}\sum_{j=1}^{X_{i}}(F_{i,j} - S_{i,j})
\end{align}
where the $F_{i,j}$ and $S_{i, j}$ are the maximal and the second maximal confidence values of the $j$th correctly classified sample of category $i$. $X_{i}$ is the number of the correctly classified sample set $\mathcal{X}_{i}$ of category $i$. $\varepsilon$ and $\rho$ are empirical parameters. 

\begin{figure*}
	\centering{\includegraphics[width=5in]{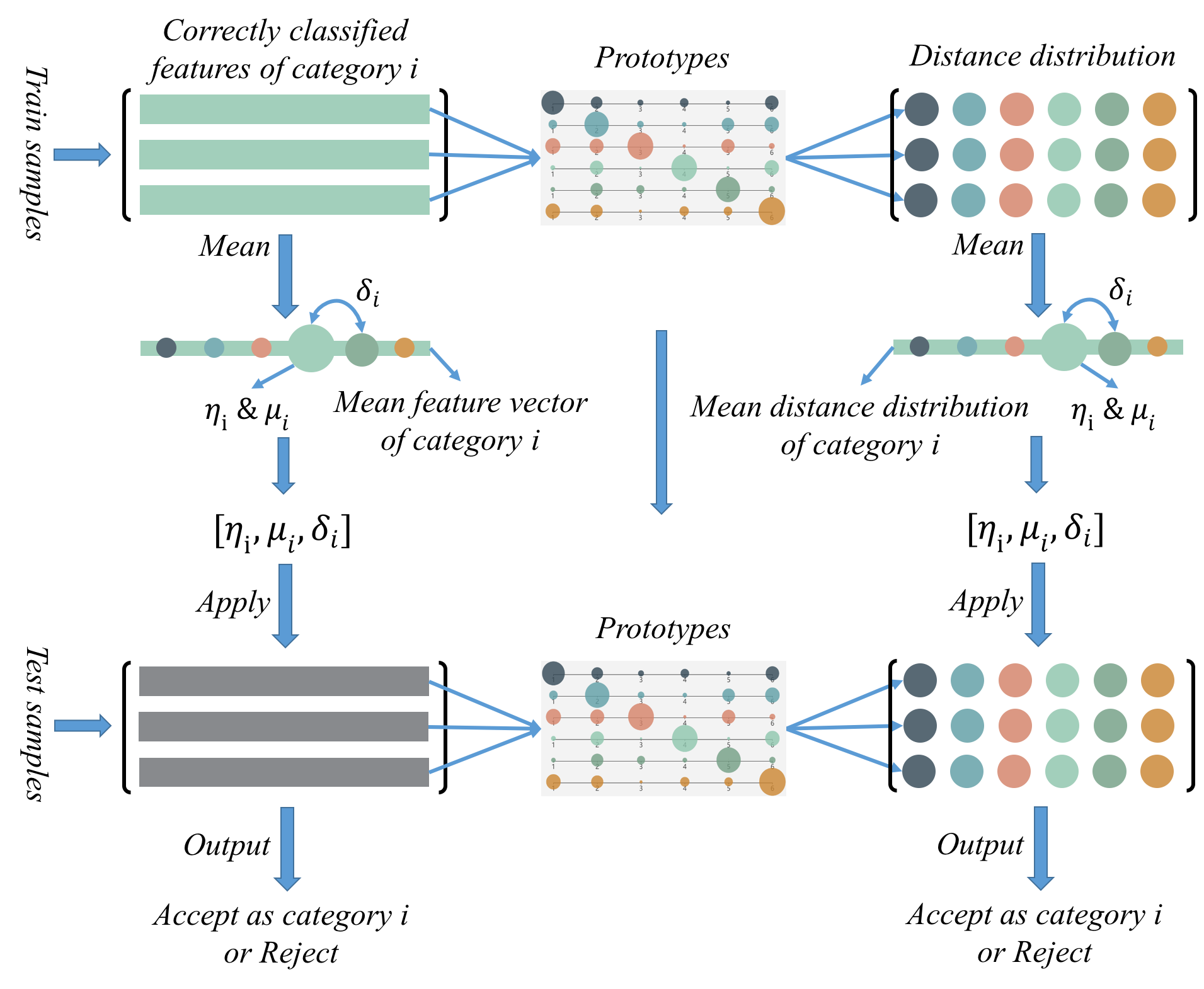}}
	\caption{The illustration of Detecting Unknowns in the P-ODN. The first column can be viewed as our previous version in \cite{shuyu2018odn}. In the P-ODN, the distance distribution matrix is calculated by using the prototypes. Thresholds calculated on the mean distance distribution are then applied on the distance distribution of the test samples to detect unknowns.}
	\label{UD}
\end{figure*}

A data sample is classified as category label \textit{l} only if the index of its top confidence value is \textit{l} and the value is greater than $\eta_{l}$. And a sample is regarded as unknowns when all of its confidence value is below $\mu$. The threshold $\delta$ is applied to help detect unknowns in hard samples, which lie between $\eta$ and $\mu$. The statistical properties of $\delta$ include correlation information between the two categories, which is a simple way of using the inter-class relation information in the activation level. If the distance is large enough, then we accept the data sample as category label \textit{l}. The process of unknowns detection is shown in the first column of Fig. \ref{UD}.

Unlike the previous version of unknowns detection in \cite{shuyu2018odn}, we modify the method to be applicable in the P-ODN framework to apply more robust unknowns detection algorithm based on the distance metric.
Fig. \ref{UD} shows the comparison of detecting unknowns between our previous work ODN \cite{shuyu2018odn} and the P-ODN. Instead of calculating the triplet thresholds based on the mean feature vectors, the distance distribution matrix of features and prototypes is calculated first. Then the triplet thresholds are calculated for each category in the same routine while based on the mean distance distribution. While the triplet thresholds are acquired, the later unknowns detection processes are similar.

The insight of this improvement is that thresholds are calculated based on statistic results of the train samples in the previous version. The statistic results, mean feature vectors, are more easily effected by the outliers. While in the P-ODN, the features are transfered into a different feature space by calculating the distance distribution matrix. Then thresholds are calculated based on the mean distance distribution. We assume that in this way the thresholds are acquired by making use of the model information, instead of only getting from the statistic information of data sets, which is more robust. Because the distance distribution can be regarded as a projection of features under the guidance of prototypes, which are trained with the model. 

\subsection*{Updating Network}\label{subsec:dwi}
After detecting the unknowns, manual labeling the unknown samples. Then these samples could be used to fine-tune the model. It has been discussed that retraining the entire system with the known data and new samples is time consuming, computational resource wasting. And it is also easy to be over-fitting, because new categories are far short of training samples. 

In our previous work \cite{shuyu2018odn}, an updating method by transferring knowledge from the trained model was proposed which helps to speed up the training stage and needs very few manually annotations. A brief retrospective of the method is given bellow.

In each iteration of the incremental training phase, a new category is incorporated to the current model, which is carried out by increasing the corresponding weight column in the classification layer of the networks. By initialization the weight column as Formula \ref{weight_init}, the knowledge of the previous model is kind of transfered to the new model.

\begin{align}\label{weight_init}
w_{N+1} = \alpha \frac{1}{N}\sum_{n=1}^{N}w_{n} + \beta \frac{1}{M}\sum_{m=1}^{M}w_{m}
\end{align}

In Formula \ref{weight_init}, the current category number is $N$ and $w_{n}$ is the weight column of the $n$th category in the classification layer of the networks. And $w_{m}$ is the weight column of $M$ most similar categories measuring by the scores of features. $\alpha$ and $\beta$ are empirical parameters.

\begin{figure*}
	\centering{\includegraphics[width=5in]{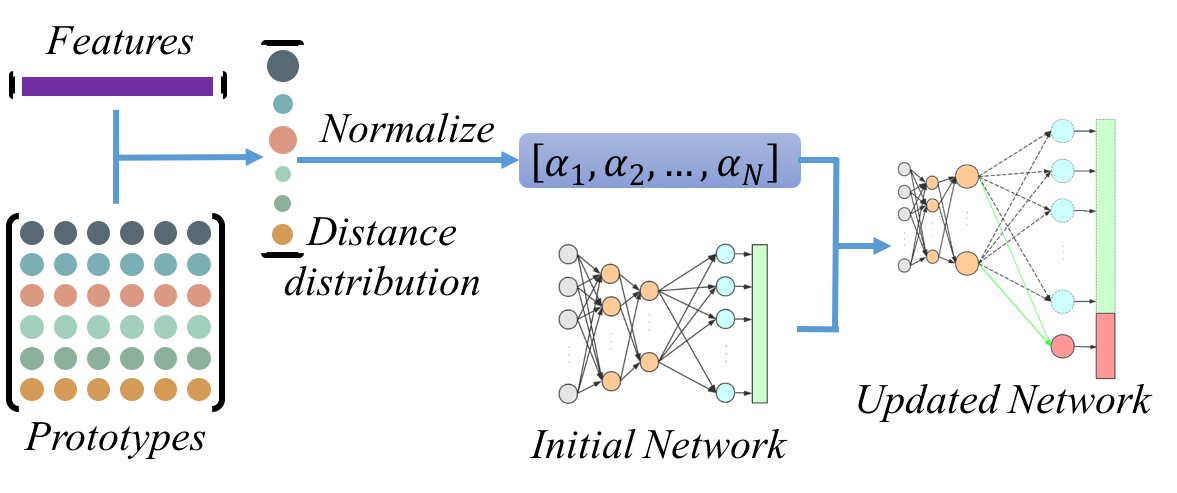}}
	\caption{The illustration of distances based weights initialization. Distance distribution is calculated between prototypes and the new sample features. Then a weight distribution can be acquired by applying the mean normalization. Finally, weights of new predictors are initialized depending on weights of the initial networks according to the weight distribution.}
	\label{DWI}
\end{figure*}

In the P-ODN, a new weight column is also increased in the classification layer to incorporate the new category. Unlike the previous version, the distance distribution of the new category sample is calculated first. Then by applying the mean normalization, we can get the distribution $[\alpha_{1}, \alpha_{2}, \cdots, \alpha_{N}]$, where $ 1 = \sum_{n=1}^{N}\alpha_{n}$, as shown in Fig. \ref{DWI}. The new weight $w_{N + 1}$ is initialized as:
\begin{equation}\label{D_weight_init}
w_{N+1} = \frac{1}{N}\sum_{n = 1}^{N}\alpha_{n}*w_{n}
\end{equation}
where $w_{n}$ is the weight column of the $n$th category, and $N$ is the current category number.

The insight of this improvement is that more robust relations of the new category and the knowns are taken into account by applying the distances based weights initialization method. By initializing the new weights like Formula \ref{D_weight_init}, the global knowledge as well as the relation knowledge can both be incorporated into the new model. First, each weight column is integrated to initialize the new weights, which guarantees new weights in the same distribution with knowns. And second, the distances based relation metric is much more robust than that in the previous work \cite{shuyu2018odn} which is measured by comparing the scores in the features.

After the networks are updated, few samples which detected in the \textit{Detecting Unknowns} module are used to fine-tune the model. As the new weights incorporate the knowledge of the previous model, the fine-tuning phase is mush less complex and very soon. We also adopt the Allometry Training method and the Balance Training method, which are proposed in \cite{shuyu2018odn}, while fine-tuning the model. Specifically, different learning rates are embedded into the classification layer to force the new weights to learn at a faster rate. And we use same few samples of each known and new category to avoid the greatly influence on the accuracy of the knowns. At the end of the Incremental training phase, the final model has the capacity to classify both knowns and unknowns.

\section*{Experiments}\label{experiments}
This section will first introduce the details of datasets and the evaluation schemes. Then, we describe the experiments setting and the exploration experiment. 


\subsection*{Datesets}
To verify the effectiveness of P-ODN, we conducted experiments on four public datasets, including UCF11 \cite{liu2009recognizing}, UCF50 \cite{reddy2013recognizing}, UCF101 \cite{soomro2012ucf101} and HMDB51 \cite{kuehne2011hmdb}. Note that we divide the datasets into knowns and unknowns to simulate the open world scenarios.

The UCF11 dataset contains $11$ action categories.For each category, the videos are grouped into $25$ groups with more than $4$ action clips in it. The video clips in the same group share some common features, such as the same actor, similar background, similar viewpoint, and so on.

The UCF50 dataset is an action recognition dataset with $50$ action categories, consisting of realistic videos taken from Youtube. For all the $50$ categories, the videos are grouped into $25$ groups, where each group consists of more than $4$ action clips. 

The UCF101 dataset is one of the most popular action recognition benchmarks. It contains 13,320 video clips from $101$ action categories and there are at least $100$ video clips for each category. 

The HMDB51 dataset is a large collection of realistic videos from various sources, including movies and web videos. It contains $6849$ clips divided into $51$ action categories, each containing a minimum of $100$ clips. 


Compared with the very large dataset used for image classification, the dataset for action recognition is relatively small. Therefor we pre-trained our model on the ImageNet dataset \cite{deng2009imagenet}.

\subsection*{Experiments setting}\label{sub:esetting}
To simulate the open world scenarios, we choose nearly half categories of each data set as knowns and the other half as unknowns, i.e. $6$ categories of UCF11 as knowns while the other $5$ as unknowns, $25$ categories of UCF50 as knowns while the other $25$ as unknowns, $50$ categories of UCF101 as knowns while the other $51$ as unknowns, and $25$ categories of HMDB51 as knowns while the other $26$ as unknowns. Then the training set of each dataset is divided into two subsets according to knowns and unknowns. The subset which contains knowns is the initial training set. A small subset is chosen from both the knowns and unknowns, which we guarantee that each category has at least $10$ samples, to form the incremental training set. Note that we use much less training samples and removing half labels of the training set to simulate the open world scenarios.

\begin{table}
	\centering
	\caption{Sample number of manually annotated unknowns needed to increase a category on average.}
	\begin{tabular}{c|c|c|c|c}
		\toprule
		\hline	
		sample number & UCF11 & UCF50 & UCF101 & HMDB51 \\ 
		\hline
		ODN & 7 & 5.8 & 5.39 & 5.57 \\
		\hline
		P-ODN & 7 & 5.4 & 5.2 & 5 \\
		\hline\bottomrule
	\end{tabular}
	\label{table:dataN}
\end{table}

After initializing from the pre-trained ImageNet model for spatial and temporal streams, we conduct the initial training phase to train prototypes, prototype radiuses of categories, and the initial model jointly. Note that both the spatial stream and the temporal stream train their own prototypes and prototype radiuses respectively. After prototypes and prototype radiuses for the two streams are trained, then triplet thresholds of both spatial and temporal streams are calculated on the initial training set based on the prototypes.

During the incremental training phase, we keep the same experiment setting as our previous work \cite{shuyu2018odn} to give a convincing comparison. Basically, we update the networks when the number of any labeled new category goes to $5$, then this category is incorporated into the current model. 

The experiments show that we use $53$ iterations (on average) to increase $51$ new categories while using dataset UCF101 ($27$ iterations for UCF50 to increase $25$, $7$ iterations for UCF11 to increase $5$ and $26$ for HMDB51 to increase $26$ new categories). So, on average, UCF101 needs to label $5.2$ ($53 \times 5 \div 51 = 5.2$) samples ($5.4$ samples for UCF50, $7$ samples for UCF11 and $5$ samples for HMDB51) for each unknown categories. A more explict comparison is shown in the Table \ref{table:dataN} between our previous ODN \cite{shuyu2018odn} and P-ODN, it is obvious that we use the same number of labeled unknown samples in the P-ODN or even less.

However, for closed set recognition, using UCF101 as an example, the training list of UCF101 split1 has $9537$ data samples of $101$ categories. On average, each category, half of the categories corresponding to the known categories and half to the unknown categories, needs $94.4$ ($9537 \div 101 = 94.4$) annotated samples. It is obvious that we use much less samples of unknowns in the open set setting.

We also conduct the closed set recognition experiments as our baseline using the same sample size as the open set setting. The results of experiments in closed set setting are much less than those of our P-ODN while both using insufficient unknown samples. So, P-ODN needs much fewer human annotations then the closed set recognition, and can achieve better performance. Worth to mention that, P-ODN suits the real world scenarios, while the closed set recognition can not handle these tasks.

\begin{figure*}
	\centering{\includegraphics[width=7in]{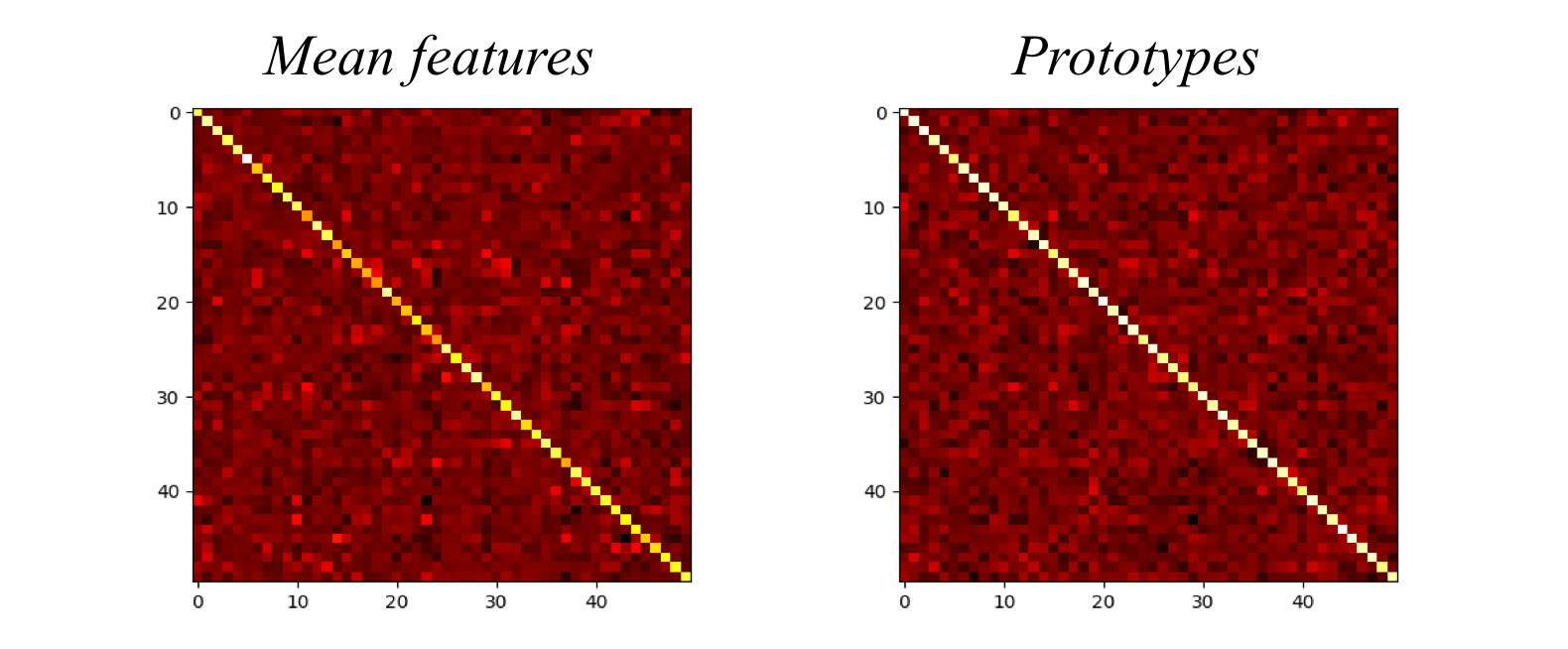}}
	\caption{Heat map of mean features and prototypes of knowns. Prototypes have much stronger response on the correctly classified values, for the diagonal line is much brighter while the upper and lower triangular matrices are much darker. }
	\label{ucf101_heat_map}
\end{figure*}

\begin{figure*}
	\centering{\includegraphics[width=6in]{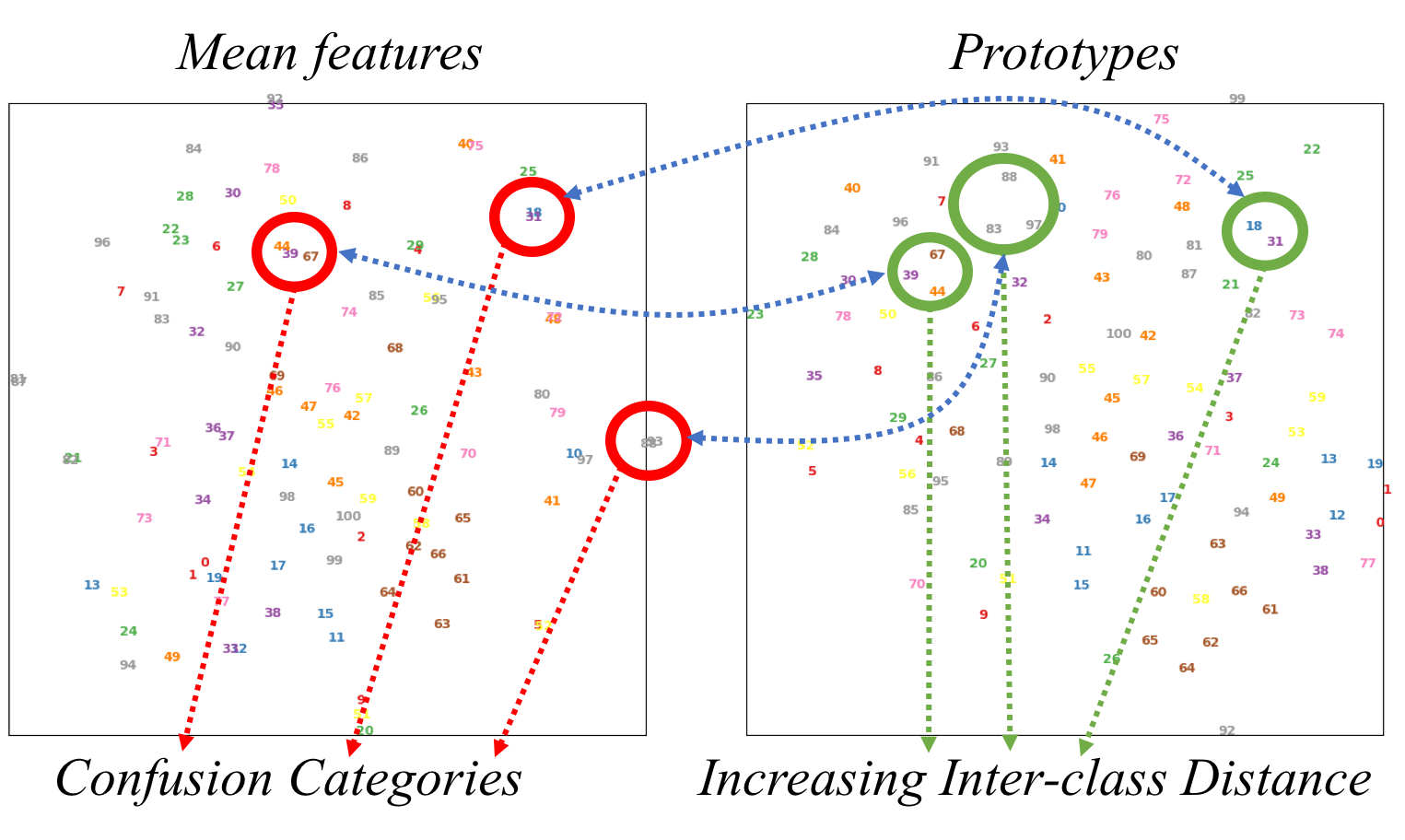}}
	\caption{T-SNE visualization of mean features and prototypes. Each colored number represents a mean feature or a prototype of a certain category. The left visualization of the mean features has more confusion categories which the inter-class distances are really short, while the prototypes can better handle the confusion categories as shown in the right sub-figure.}
	\label{ucf101_tsne_mav_prototypes}
\end{figure*}

We use the Tensorflow toolbox \cite{abadi2016tensorflow}. And we give the results on Inception-resnet-v2. The network weights are trained using the mini-batch stochastic gradient descent with momentum (set to $0.9$). We resize all input images to $340 \times 256$, and then use the fixed- crop strategy \cite{wang2015towards} to crop a $299 \times 299$ region from images or their horizontal flip. 

\subsection*{Exploration experiments}\label{subsec:ee}

\textbf{Benefits from prototypes}. To illustrate the improvement of prototypes, we firstly conduct an exploration experiments on UCF101 with GoogLeNet \cite{szegedy2015going}. 

We visualize the heat map of the mean features and prototypes of the knowns, as shown in Fig. \ref{ucf101_heat_map}. We can see that prototypes have much stronger response on the correctly classified values, for the diagonal line is much brighter while the upper and lower triangular matrices are much darker. 

As shown in Fig. \ref{ucf101_tsne_mav_prototypes}, we reduce dimensions of the mean features and the prototypes, and visualize them by applying t-SNE. In the figure, each colored number represents one mean feature of a certain category or one prototype of a certain category. The left visualization of the mean features has more confusion categories which the inter-class distances are really short, while the prototypes can better handle the confusion categories as shown in the right sub-figure.

The comparison shows that the prototypes are more suitable for representing category centers. Because the prototypes have stronger response on the collected classified values and longer inter-class distances. The two advantages help to complement more robust unknowns detection and guide much better features training.

\section*{Results and analysis}

In this section, we report the experimental results and give the analysis of results.

\subsection*{Evaluation of detecting unknowns}\label{results}
%
%

\begin{table}
	\centering
	\caption{Unknowns detection results of P-ODN.}
	\begin{tabular}{c|c|c|c|c}
		\toprule
		\hline	
		F-score & UCF11 & UCF50 & UCF101 & HMDB51 \\ 
		\hline
		OSDN\cite{bendale2016towards} & 82.59\% & 75.34\% & 72.1\% & 50.31\% \\
		\hline
		ODN\cite{shuyu2018odn} & 87.39\% & 74.91\% & 73.35\% & 63.70\% \\
		\hline
		P-ODN & 89.12\% & 80.14\% & 75.45\% & 66.79\% \\
		
		P-ODN + radius & \textbf{89.50\%} & \textbf{82.15\%} & \textbf{76.2\%} & \textbf{67.36\%} \\
		\hline\bottomrule
	\end{tabular}
	\label{table:fscore}
\end{table}

In this subsection, we aim to evaluate the unknowns detection performance of our P-ODN on UCF11, UCF50, UCF101 and HMDB51. As mentioned before, we conduct this evaluation at the end of initial training phase as \textit{Evaluation phase 1}. The experimental results are summarized in Table \ref{table:fscore}. The first row of the results is the performance of OSDN proposed in \cite{bendale2016towards}, we conduct the method on the action recognition task. The second row is the performance of our previous work \cite{shuyu2018odn}. And the third row is the performance of P-ODN with \textit{prototype module} only, the last row is P-ODN with both \textit{prototype module} and \textit{prototype radius module}. We can see P-ODN with both \textit{prototype module} and the \textit{prototype radius module} improves the ODN by $2.11\%$ on UCF11, $7.24\%$ on UCF50, $2.85\%$ on UCF101 and $3.66\%$ on HMDB51.

\begin{table}
	\centering
	\caption{Recognition results of P-ODN.}
	\begin{tabular}{c|c|c|c|c}
		\toprule
		\hline	
		TOP1 Acc. & UCF11 & UCF50 & UCF101 & HMDB51 \\
		\hline
		baseline & 85.1\% & 84.95\% & 72.01\% & 44.58\% \\ 
		\hline
		ODN \cite{shuyu2018odn} & 94.91\% & 93.73\% & 76.07\% & 46.01\% \\
		\hline
		P-ODN & 94.9\% & 95.16\% & 77.21\% & 47.84\% \\
		
		P-ODN + radius & \textbf{95.31\%} & \textbf{96.15\%} & \textbf{78.64\%} & \textbf{49.09\%} \\
		\hline\bottomrule
	\end{tabular}
	\label{table:accuracy}
\end{table}

Unknowns detection based on the prototypes are more robust. First the prototypes are more discriminable than mean features. Second the prototypes can guide the features to be trained better, which helps to improve the intra-class compactness and inter-class distance of the feature representation. We also learn the triplet thresholds based on the prototypes which would contain the knowledge of the model itself. Then the much discriminable features and the model based triplet thresholds both lead to a great improvement on the performance of unknowns detection.

\subsection*{Evaluation of classification on both knowns and unknowns}
In this subsection, we aim to evaluate the classification performance of our P-ODN on UCF11, UCF50, UCF101 and HMDB51. We conduct this evaluation at the end of incremental training phase as \textit{Evaluation phase 2}, the final classification accuracy of both knowns and unknowns is viewed as the most important performance indicator of open set recognition tasks. The experimental results are summarized in Table \ref{table:accuracy}. First, we carry out the closed set recognition experiments while using the same quantity of samples as those of our open set setting. Under the closed set setting, all training samples should have labels, so we provide labels of both knowns and unknowns. The result is shown in the first row as our baseline. The rest of Table \ref{table:accuracy} are results under the open set setting. The second row is the result in our previous work \cite{shuyu2018odn}, and we add the experiment on UCF11 here, since we did not use UCF11 in the previous work. The last row is P-ODN with both \textit{prototype module} and \textit{prototype radius module}, which achieves the best performance. We can see P-ODN finally improves the ODN by $0.3\%$ on UCF11, $2.42\%$ on UCF50, $2.57\%$ on UCF101 and $3.08\%$ on HMDB51. And  furthermore, P-ODN finally improves the baseline by  $10.21\%$ on UCF11, $11.2\%$ on UCF50, $6.63\%$ on UCF101 and $4.51\%$ on HMDB51.
\begin{figure*}
	\centering{\includegraphics[width=7in]{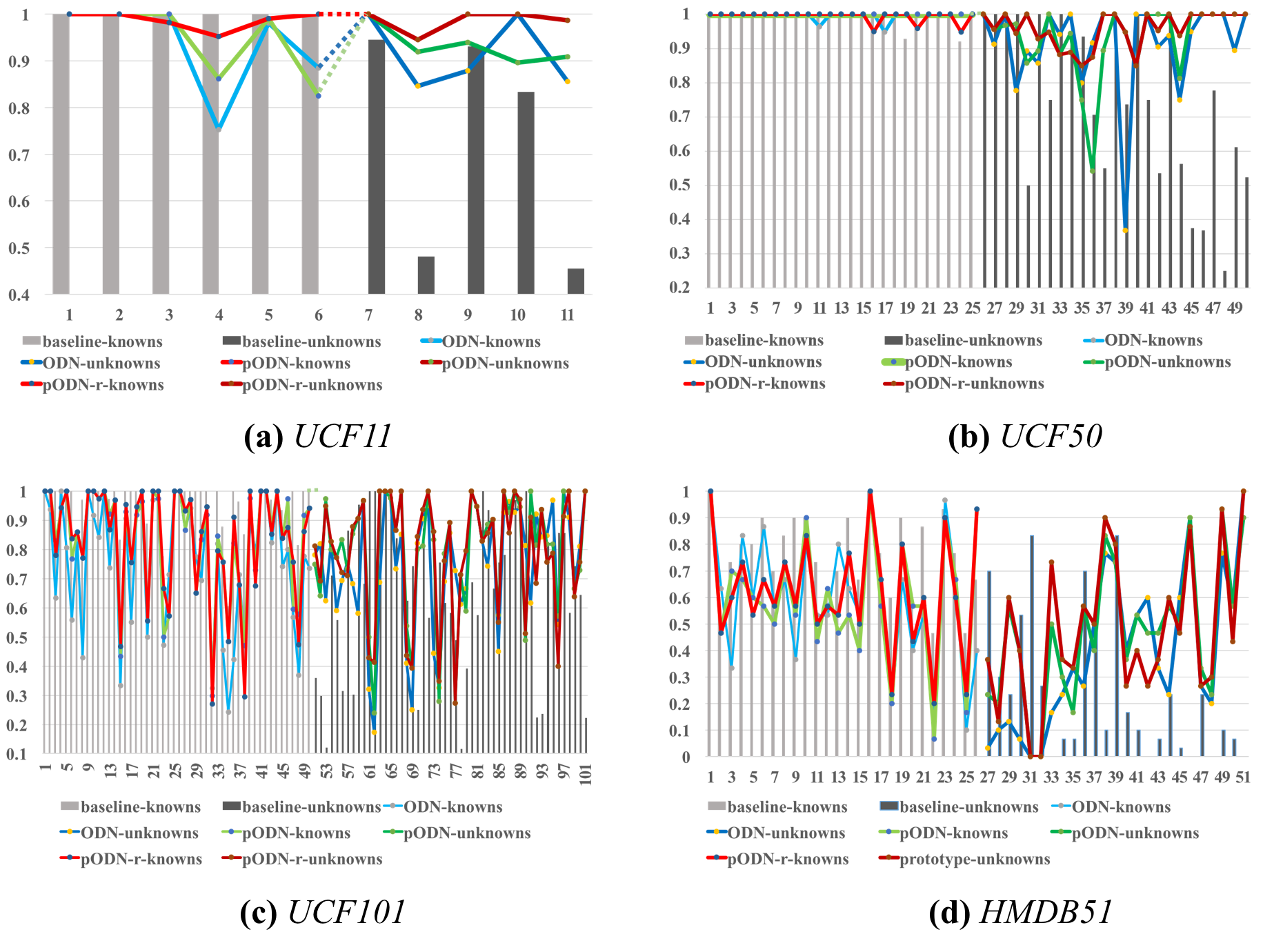}}
	\caption{Comparison of baseline method, ODN \cite{shuyu2018odn}, P-ODN and P-ODN with radius on category accuracy of UCF11, UCF50, UCF101 and HMDB51 in real world scenarios. Each sub-figure is corresponding to a data set, taking the sub-figure \title{(a)} as an example. The light gray bar denotes the accuracy of knowns of baseline, while the dull gray denotes the accuracy of unknowns of baseline. The light blue line denotes the accuracy of knowns of ODN, while the dark blue denotes the accuracy of unknowns of ODN. And so on, the light green denotes the knowns of P-ODN and dark green denotes the unknowns of P-ODN. The light red and the dark red denote the knowns and unknowns of P-ODN with radius respectively.}
	\label{fig:subfig}
\end{figure*}
A more explicit illustration can be see in Fig. \ref{fig:subfig}. Each sub-figure is corresponding to a data set, UCF11, UCF50, UCF101 and HMDB51. Take the sub-figure \ref{fig:subfig:a} as an example, we compare the accuracy of both knowns and unknowns on UCF11 with four methods, which are baseline, ODN \cite{shuyu2018odn}, P-ODN and P-ODN with radius (P-ODN with both \textit{prototype module} and the \textit{prototype radius module}). The light gray bar denotes the accuracy of knowns of baseline, while the dull gray denotes the accuracy of unknowns of baseline. The light blue line denotes the accuracy of knowns of ODN, while the dark blue denotes the accuracy of unknowns of ODN. And so on, the light green denotes the knowns of P-ODN and dark green denotes the unknowns of P-ODN. The light red and the dark red denote the knowns and unknowns of P-ODN with radius respectively.

We can see that while using the baseline method, the knowns which are trained with abundant data samples can achieve a much better performance than the unknowns which are trained with insufficient data samples with labels. Our methods can improve greatly on unknowns while using insufficient samples. Though, the performance on the knowns may decrease slightly, since the fine-tuning phase incorporates new data continuously. In Fig. \ref{fig:subfig}, we can see P-ODN with radius is generally above the other methods, which achieves the best performance. Note that, different from the baseline method which is provided with all labels beforehand, the other three methods should detect the unknowns first, then manual labeling the unknowns. Therefore, open set recognition is more realistic then the closed set setting. 

\section*{Conclusion}

This paper proposed a prototype based Open Deep Network (P-ODN) for open set recognition. We introduce prototype learning into open set recognition tasks by training prototypes of categories and prototype radiuses with a \textit{prototype module} and a \textit{prototype radius module}. Then a distance metric method is applied to detect unknowns, which is based on the prototypes and more robust. In the incremental training phase, a distances based weights initialization method is employed to fast acquire the knowledge of model and speed up the fine-tuning process. Experimental results show that, our P-ODN can effectively detect and recognize new categories with little human intervention and achieve state-of-the-art performance on UCF11, UCF50, UCF101 and HMDB51 datasets.

In this paper, we have proved the importance of more discriminable centers (or prototypes) on the open set recognition tasks. More characteristic features which have larger margin among categories will further improve the performance of unknowns detection. In addition, method \cite{ge2017generative} utilizes GAN to generate unknown samples and uses them to train the neural networks also has potential to improve the recognition performance of unknowns. In the future work, we will conduct more experiments as mentioned above to further improve the performance of open set recognition.

\bibliography{sample}
%
%

\end{document}